\documentstyle[12pt,named]{article}

\newcommand{\lingform}[1]{{\it #1}}

\title{NLG vs. Templates\thanks{This research has been partially supported
by Rome Laboratory (US Air Force) under contract F30602-94-C-012.}}
\author{Ehud Reiter\thanks{After 1 August 1995, Dr. Reiter's address will be
Department of Computing Science, University of Aberdeen,
King's College, Aberdeen AB9 2UE, BRITAIN.
His email address will be {\tt ereiter@csd.abdn.ac.uk}}\\
CoGenTex, Inc\\
840 Hanshaw Rd\\
Ithaca, NY 14850 USA\\
email: {\tt ehud@cogentex.com}}

\date{ }

\begin{document}

\pagestyle{empty}
\bibliographystyle{named}
\setlength{\baselineskip}{13pt}               

\maketitle

\begin{abstract}
One of the most important questions in applied NLG is what benefits
(or `value-added', in business-speak) NLG technology offers over
template-based approaches.  Despite the importance of this question
to the applied NLG community, however, it has not been discussed
much in the research NLG community, which I think is a pity.  In this paper,
I try to summarize the issues involved and recap current thinking on this
topic.  My goal is not to answer this question (I don't think we know enough
to be able to do so), but rather to increase the visibility of this issue
in the research community, in the hope of getting some input and ideas
on this very important question.  I conclude with a list of specific
research areas I would like to see more work in, because I think they would
increase the `value-added' of NLG over templates.
\end{abstract}

\section{Introduction}

There are thousands, if not millions, of application programs in
everyday use that automatically generate texts; but probably fewer than
ten of these programs use the linguistic and knowledge-based techniques that
have been studied by the Natural-Language Generation (NLG) community.
The other 99.9\% of systems use programs that simply manipulate character
strings, in a way that uses little, if any, linguistic knowledge.
For lack of a better name, I will call this the `template' approach.

In order for NLG technology to make it out of the lab and into everyday
fielded application systems, the NLG community will need to prove
that there are at least some niches
where using linguistic/AI approaches in a text-generation system
provides real commercial advantages, such as reducing the effort required
to build (or maintain) the system, or improving the effectiveness of the
generated texts.
Determining under what conditions and in what aspects NLG techniques
are `better' than character-string
manipulation is of utmost importance to the applied NLG community, and
should also be of interest to the research community;
if nothing else,
research funding for NLG is likely to increase if there are
a large number of successful
fielded systems that use NLG technology.

In this paper, I will use the term {\em automatic text generation (ATG)}
to refer to any computer program
that automatically produces texts from some input
data, regardless of whether NLG or template technology is used internally.
The topic of this paper is thus when is NLG `appropriate technology'
for building ATG systems, and when should simpler approaches be used.
My goal is not to provide a definitive
answer to this question, because I don't think we (currently) know enough to be
able to do this, but rather to present the issues, summarize
comments made by other people, and present some opinions of my own.  Hopefully
this will help start a discussion within the community about this very
important but (so far) somewhat ignored issue.

\section{Template systems}
\label{template}

All ATG systems are, of course, simply computer programs that run on some
input data and produce an output (the text) from this data.
Non-linguistic (`template') text-generation is done via manipulating
character strings; the user writes
a program which includes statements such as `include \lingform{XXXXX}
if condition Y is true, and \lingform{YYYYY} otherwise.'  This program
can be written directly in a programming language such as Lisp or C++,
or it can be specified via a `mail-merge' system which allows conditional
texts (eg, Microsoft Word).  The key difference between
this approach and NLG is that all manipulation is done at the
character string level; there is no attempt to represent the text in any
deeper way, at either the syntactic or `text-planning' level.

To the best of my knowledge, most programming languages and mail-merge
environments provide very little, if any, support for manipulating texts
in even the simplest `linguistic' manner.
For programming languages, the most sophisticated
feature that I am aware of is the \verb+~P+ construct in the Lisp
\verb+format+ function, which will do some simple pluralizations
(eg, \lingform{win} vs. \lingform{wins},
or \lingform{try} vs. \lingform{tries}) depending on whether a numeric
parameter is one or not.

Mail-merge systems can have slightly more sophisticated capabilities, such as
automatically capitalizing
an inserted word if it is the first word of a sentence.  However, even
something as simple
as changing pronouns according to gender
needs to be explicitly programmed.
Some mail-merge systems are integrated with grammar checkers that might in
theory be able to handle some low-level syntactic problems such as
verb agreement, \lingform{a} vs. \lingform{an}, and elimination of multiple
commas; however, current grammar checkers may not be robust enough to be able
to do this in a reliable fashion.

\subsection{Example: Apple Balloon Help}

A simple example of template-based generation is
the Apple Macintosh Balloon Help system.  It can produce texts such as
\begin{quote}
This is the kind of item displayed at left.  This shows that test data
is a(n) Microsoft Word document.
\end{quote}
and
\begin{quote}
This is a folder --- a place to store related files.  Folders can contain
files and other folders.

The icon is dimmed because the folder is open.
\end{quote}
In the first text, {\em test data} and {\em Microsoft Word} were inserted into
template slots for `filename' and `application program.'  Note the use of
\lingform{a(n)}; even this simple type of agreement is not done in the
Balloon Help system.
In the second
text, the last sentence ({\em The icon is dimmed because the folder is open})
only appears when the mouse is positioned over an open folder; just the
first two sentences will appear if the mouse is positioned
over a closed folder.  This is an example of conditional text.

\subsection{Example: Employee Appraiser and Performance Now}
\label{performance now}

Two more sophisticated `non-linguistic' automatic text-generation systems are
Austin-Haynes {\em Employee Appraiser} and
KnowledgePoint's {\em Performance Now}.  Both of these systems help managers
write appraisals of employees (eg, for justifying salary increases).
Each system provides a set of general evaluation topics, such
as {\em Communication}, which are broken up into more specific subtopics,
such as {\em Communicates ideas verbally}.  Managers give employees rankings
on each of these subtopics, and the system then composes a complete appraisal,
which the manager can edit in a word processor.

In Employee Appraiser, when the manager chooses a subtopic rating, the system
retrieves an appropriate paragraph from a library and does some simple
linguistic processing.  In particular, the manager can specify whether the
employee is male or female, and whether the report should be written in
second or third person.
This affects
the pronouns used in the text (eg, \lingform{he}, \lingform{she}, or
\lingform{you}), and also verb conjugation (eg, \lingform{you do} vs.
\lingform{he does}).  This is the most sophisticated syntactic processing
that I am aware of in a `template' system.

Performance Now does less syntactic processing (it does not allow the manager
to choose between second and third person; only third-person is possible),
but it does do some simple sentence planning.  In particular,
Performance Now combines all the information about a particular
high-level topic (such as {\em Communication}) into a single
paragraph, and this requires the system to use conjunctions and pronouns,
and to add initial conjuncts (eg, \lingform{Furthermore}) to sentences.

An example output of Performance Now is
\begin{quote}
Bert does not display the verbal communication skills required, and his
written communications fall short of the quality needed.  Additionally,
he does not exhibit the listening and comprehension skills necessary
for satisfactory performance of his job.
\end{quote}
The system has composed this from three separate phrases retrieved from its
library.  The first two phrases are combined with \lingform{and} to produce
the first sentence above.  The third phrase is left as a separate sentence,
but the conjunct \lingform{Additionally} is added to it, and the subject is
pronominalized.  The system also orders phrases by putting the most
positive ones first, and most negative ones last (this is not shown in the
above example).

Performance Now performs the most sophisticated sentence-planning of any
`template' system that I am aware of; indeed, one might argue that Performance
Now is doing enough linguistic processing that it really should be regarded as
a (simple) NLG system.  KnowledgePoint's marketing literature in fact
stresses their `IntelliText$^{TM}$' technology, which ``generates clear,
logical sentences and modifies them to work together as if you wrote them
yourself''.  This is the only mass-market system I am aware of which advertises
NLG-like abilities as part of its competitive advantage.

\newpage

\section{Advantages of NLG}

Many advantages of NLG over templates have been described in the literature.
In this section, I try to summarize these arguments, paying special
attention to those arguments
that seem important to the success of current applied NLG systems.

\subsection{Maintainability}

One reason for using NLG is {\em maintainability}; template-based generators
can be difficult to modify according to changing user needs.
This has been a real factor in the success of the FoG weather-report
generation system, for example.  To quote \cite[page 53]{goldberg:ieee94}
\begin{quote}
Experience has shown that [template-based weather report generators are]
difficult to maintain.  This has hampered the testing and implementation
of the software and has made it difficult to update the program for changing
user requirements.  This is a critical factor.  Although the Canadian
textual forecast products fall into several common broad categories (marine
forecasts, public forecasts, and so on), each category contains many
regional variations.  Also, content, structure, and terminology tend to vary
with time, albeit slowly.  To succeed, a system must address variations
between forecast types, variations between geographical regions in a forecast
type, and gradually changing requirements.
\end{quote}
Making even a slight-change to the output of a template-based generator
may require a large amount of recoding (of programs) and rewriting
(of templates); in contrast, such a change may be straightforward to make in
linguistically-based system.  To take one simple example,
if a user wishes to change a text-generation system so that dates are
always at the beginning of sentences (eg,
\lingform{In 1995, a severe winter is expected\/} instead of
\lingform{A severe winter is expected in 1995}), this can easily be done
with almost any linguistically-based NLG system.  With a template system,
in contrast, making this change may require rewriting a large number of
template fragments.\footnote{This assumes that the system
uses a large number of templates.  If only a small set of templates is needed
to generate the system's texts, maintaining them is unlikely to be a problem.}

There is an interesting analogy with expert systems here.  Early expert
systems, such as the R1/XCON system used to configure computers at Digital
Equipment Corporation \cite{xcon:aij}, were partially justified on the grounds
that they were easier to maintain and modify than `conventional' programs
that performed the same task.  It subsequently became clear, however
(eg, \cite{soloway:aaai87}), that maintaining expert systems, although still
perhaps easier than maintaining conventional programs that performed similar
tasks, was not as simple as the initial enthusiasts had thought it would be.
We as yet have little data on how easy/difficult it really is to maintain
fielded NLG systems; \cite{kittredge:anlp94} is the only paper I am aware
of that discusses the maintenance of fielded NLG systems.

\subsection{Improved Text Quality}

Another advantage of NLG-based systems is that they can produce
{\em higher-quality output}.  It is useful when
discussing output quality to distinguish between aspects of quality that
arise from the three different processing
stages used in most applied NLG systems \cite{reiter:arch}:
content/text planning, sentence planning, and syntactic realization.

\subsubsection{Content Planning}

The ability to vary the {\em information content} of a text in a fine-grained
and flexible way may be the most important `quality' enhancement
of all; it allows NLG-based systems to include whatever information is deemed
important in a text, and leave out unimportant information.  This has
been especially important in letter-generation systems
(eg, \cite{springer:iaai,coch:anlp94}) which have been
one of the most popular NLG applications to date.
To quote \cite[page 68]{springer:iaai}
\begin{quote}
An automated form-letter system originally formed the core of this
organization's correspondence facility.  Because its letters must specifically
discuss different kinds of financial transactions, the system has grown
to include close to 1,000 different form letters to address the simplest
divisions of common problems.  However, in practice most of these letters
are never used: Customer service representatives, working under pressure
to handle as many cases as quickly as possible, tend to use 10 to 20
letters that are close enough to describing the client's situation rather
than take the time to discriminate between slight variations within the
form library.  When a client's situation even slightly varies from these
forms or encompasses a combination of topics addressed in separate form
letters, a new letter must be composed by hand if the client is to be
convinced that s/he has received individual attention.  Form-letter systems
might come cheap, but they don't always stay that way, and the quality of
output for any particular situation can never be very high.
\end{quote}
In other words, if a text-generation system has to be able to generate texts
that are appropriate for many different kinds of situations, it may be
difficult to use as well as build; and there may be a strong argument for
building a system which uses knowledge-based techniques to represent the
desired content of the output, and then generates an appropriate textual
presentation of this context.

\subsubsection{Sentence Planning}
\label{aggregation}

Most applied NLG systems have a sentence planning module that handles
aggregation, referring-expression generation, sentence formation, and
lexicalization \cite{reiter:arch}.
Performing these tasks well can greatly enhance
the readability of a text.  Consider, for example, the difference between
\begin{quote}
The house is white.  The house is large.  The house is owned by
John.  The house is on Sullivan Street.  The house is next to the elementary
school
\end{quote}
and
\begin{quote}
John owns a large white house on Sullivan Street.
It is next to the school.
\end{quote}
This is an example of aggregation \cite{dalianis-hovy:nlgw93}.
Aggregation is mentioned as one of the most important benefits
provided by the PLANDOC system \cite{mckeown:anlp94}.
PLANDOC summarizes the history
of an engineer using a simulation package, and generates text such as
\begin{quote}
This refinement activated  DLC for CSA 2111 in 1995 Q3, for CSAs 2112 and 2113
in 1995 Q4, and for CSAs 2114, 2115, and 2116 in 1996 Q1.
\end{quote}
If each of the activations was expressed by a separate sentence, the above
message would require six separate sentences, and would be much longer.

An interesting open question is how much sentence planning can be done
without having a `proper' syntactic representation of the text.
The Performance Now (Section~\ref{performance now}) system demonstrates
that some simple aggregation can be done even if phrases are represented
as character strings,
but it seems doubtful whether more sophisticated aggregation (eg, ellipsis
or relative-clause introduction) is possible
without a syntactic representation of phrases.

\subsubsection{Syntactic Realization}

Texts that are comprehensible but ungrammatical can be
annoying to readers, and it may be expensive (in terms of programming effort)
to set up a template system to correctly handle agreement, morphology,
punctuation reduction, and other `low-level' phenomena.
It is straightforward, in contrast, for an NLG system to handle
such phenomena.

Interestingly enough, however, I am not aware of any applied NLG
system whose success is primarily based on better syntactic (or morphological)
processing.  Systems such as FoG \cite{goldberg:ieee94} and PLANDOC
\cite{mckeown:anlp94} do possess sophisticated syntactic realization systems,
but I do not believe that their success derives from the fact that they
can get agreement or morphology right.  Template systems such as
Employee Appraiser are, after all, able to handle some agreement phenomena.
Particularly if a developer is only concerned with relatively
straightforward phenomena (eg, noun pluralization, or noun-verb agreement), it
may be easier for him or her to `hack' something together that
appropriately manipulates character strings, instead of trying to build
explicit syntactic structures that can be processed by an NLG system.

Also, in many cases texts can be phrased in a manner which minimizes the
need for syntactic adjustment.  For example, problems will occur with the
template \lingform{N iterations were run}  when N is 1; these problems can
be avoided, however, by changing the text to \lingform{Number of iterations
run: N}.

A good syntactic module may of course be needed to support a sophisticated
content determination or sentence planning module.  For example, as mentioned
in the previous section, proper syntactic representation is probably needed
for many kinds of aggregation.
By itself, however, good syntactic processing
may not provide much `competitive advantage' to an NLG system.

\subsection{Other advantages}
\label{standards}
\label{multilingual}

Two other advantages of NLG that may be important in some cases are
\lingform{multilingual output} and \lingform{guaranteed conformance
to standards}.  Multilingual output can of course be achieved with templates;
many error-message systems, for example, are localized to other languages
simply by inserting a new set of format strings.  The quality of texts
generated by this approach is not high, but this may be acceptable in some
circumstances.

At the other extreme, multilingual output could also be achieved by building
several separate systems, one for each target language.  Such a system
would be expensive to construct and might prove difficult to maintain, however.

The FoG weather-report generation system \cite{goldberg:ieee94}
probably owes some of its success
to the fact that it can produce texts in both French and English.  Weather
reports in Canada must be produced in both French and English, and if reports
are first written in one language and than translated into the other, there
may be a significant delay before the translated reports are available
(even a one-hour delay can be significant for a 24-hour weather forecast).
The FoG system enables the forecaster to simultaneously produce both English
and French versions of the forecast, thus eliminating
this delay.\footnote{Also, a forecaster who uses FoG is not dependent on a
third-party to translate his or her forecasts; this
feeling of `more control' may be a significant plus to some users.}

The final advantage of NLG I'd like to mention is guaranteed conformance
to document standards, including writing standards such as AECMA Simplified
English \cite{simplified-english:book},
and content standards such as DoD-2167A \cite{2167A}.
In many domains it is essential that documents conform to such standards,
and rules such as `sentences should not be longer than 20 words' 
or `sentences should not contain more than three sequential nouns' (from AECMA
Simplified English) may be easier to enforce in an NLG system, which
can paraphrase or reword texts to meet such constraints.
An NLG system could,
for example, take sentence-length constraints into account when
making aggregation decisions.
I do not know
of any current applied NLG system for which standards conformance is an issue,
but this may become important in future applications.

\section{Advantages of Templates}

Templates, of course, also have advantages over NLG.  The most basic of these
is probably that NLG systems
cannot generate text unless they have a representation of the information
that the text is supposed to communicate; and in the great majority of
today's application systems, such representations do not exist.

For instance, suppose a scientific program wishes to inform the user that
N iterations of an algorithm were performed.
In principle, NLG techniques could
be used to improve the handling of special cases such as N=0 and N=1, so that
the system could produce
\begin{quote}
No iterations were performed\\
1 iteration was performed\\
2 iterations were performed
\end{quote}
However, doing this with NLG techniques
would require the system to have either a declarative
representation of concepts such as algorithms and iterations; or a syntactic
representation of the sentence \lingform{N iterations were performed}.
Neither of these is likely to exist in a scientific program, and few
scientific programmers would bother putting them in.  Instead, such programmers
would either accept low-level syntactic problems (eg, the output
\lingform{1 iteration were performed});
use an alternate formulation
that did not suffer from this problem (eg, \lingform{number of iterations
performed: 1}); or write special code to produce the appropriate output
when N is 0 or 1.

This is perhaps an extreme case, but it illustrates the point that switching
to NLG will be expensive if the application does not already have a
declarative domain knowledge base and/or syntactic representations of output
text, and no one is going to pay this cost if the resultant improvement in
text quality (or system maintainability) is not perceived as significantly
enhancing the usefulness of the application.  Since knowledge-based
application systems are still rare, and even the ones that do exist often
do not have all the information that an NLG system would need, it may be
the case that NLG is not the most
appropriate technology for many current text-generation applications.

Besides the above problem, NLG also suffers from generic problems that
are common to all new technologies.
There are very few people who can build NLG systems,
compared to the millions of programmers who can build template systems;
there is also very little awareness of what NLG can (and cannot) do among
most developers of application systems.
Additionally, there is very little in the
way of reusable NLG resources (software, grammars, lexicons, etc), which
means that most NLG developers still have to more or less start from scratch.
Finally, the fact that NLG is an experimental technology means that
conservative developers may want to avoid using it.  As mentioned above,
these problems are common to all new technologies, and will evaporate
with time if NLG proves to be a truly useful and valuable technology.

\section{Hybrid Systems}
\label{hybrid}

It is of course possible to build ATG systems that use both NLG and template
techniques.  To date, two variants of this have been particularly common:
\begin{itemize}
\item Systems that embed NLG-generated fragments into a template slot,
or that insert canned phrases into an NLG-generated matrix sentence.
The IDAS system \cite{idas:aai}, for example,
could insert generated referring expressions
into templates such as \lingform{Carefully remove the wrapping
material from X}, and also could insert canned phrasal modifiers such as
\lingform{using both hands} into an otherwise generated sentence.
\item Systems that use NLG techniques for `high-level' operations such as
content planning, but templates for low-level realization
(eg, \cite{pitt:medical}).
\end{itemize}
The basic goal of such systems is to use NLG where it
really `adds value', and to use simpler approaches where NLG is not needed
or would be too expensive.  The decision on where NLG should be used can
be based on cost-benefit analysis \cite{idas:ijcai}.

Real-world decisions about where NLG should be used in a hybrid system
are likely to be based on practical criteria as well as theoretical ones.
NLG modules that are slow, error-prone, written in unusual programming
languages, and/or difficult to maintain will
of course not get used much in real applications.  But also, even a
well-engineered module is unlikely to get used if it does not fit into
the way the developer wishes to build his or her system, or give the developer
sufficient control over the system's output.  For example, a system that
reserves the right to reorder sentences based on some rhetorical model
may be unacceptable to a developer who insists that the sentences must appear
in a specific order that he or she thinks is best.

Another way of saying this is that NLG shouldn't `get in the way'.
Developers will use NLG modules and techniques if NLG helps them produce the
kind of texts they want to produce; if an NLG system is seen not as a helpful
tool but as something that needs to be worked around, it will not be used.
NLG should also only be used when it clearly increases maintainability, text
readability, or some other important attribute of the target application
system.  If a certain portion of the output text never varies, for example,
it would be silly to generate it with NLG, and much more sensible to
simply use canned text for this portion of the document.

I believe, by the way, that most current hybrid systems
use `real NLG' in content-determination and perhaps sentence-planning,
and use template techniques mainly in syntactic realization.  This may
simply be a coincidence, but it may also suggest
that much of the real `value-added' of many NLG systems may be
in the high-level processing, not in ensuring correct syntax.

\section{Making NLG More Useful}

There are several areas where I think more academic research could help
improve the advantages of NLG-based text
generators over template-based systems.\footnote{This list has been heavily
influenced by discussions with Chris Mellish.}

\begin{description}
\item[Aggregation:] As mentioned in Section~\ref{aggregation},
aggregation (eg, clause
combining, ellipsis, conjunction reduction) is something that
seems to significantly add to text quality in many
circumstances.  Yet, there has been surprisingly little research on this
very important and interesting topic.
\item[Standards Conformance:] Being able to generate text that is guaranteed
to conform to a writing or content standard could be a big selling point of
NLG in some circumstances (Section~\ref{standards}).
Currently, however, it is only possible to
enforce `low-level' syntactic and lexical standards; I think it would
be very interesting to examine how higher level standards, such as
`only one topic per paragraph', might be enforced.
\item[Multilinguality:] Multilingual output is another feature that could be
a strong selling point for NLG in many circumstances
(Section~\ref{multilingual}).  But although many
multilingual NLG systems have been built, surprisingly little research has
been done on the principles underlying multilinguality.
For example, where can language
independent modules be used in a multilingual system, and where is this
impossible?
\item[Multimodality:] Real-world documents include diagrams, tables, and other
graphics as well as text;
and real-world documents also use {\em visual formatting},
such as font changes and bulletized lists, within texts.  Additionally,
on-line documents often include hypertext links.  A system that generates
documents is going to be much more useful if it can
combine text and graphics, use appropriate visual formatting, and insert
hypertext links into online documents.
\item[Hybrid Systems:] It seems safe to predict that many fielded NLG systems,
at least in the near term, will use a hybrid approach, ie, they will use both
template and NLG technology.  But little is known about how this should best
be done; if we want to insert a generated referring expression into a template
slot, for example, what constraints does the template need to satisfy in order
for this to produce correct output?
And how should `templates' used by an
NLG system be authored; can we develop a nice authoring environment which
enforces any necessary constraints in an intuitive manner?
\item[Modifying Generated Text:]
All real-world NLG systems that I am aware of allow the human user to
modify the generated texts.  But there are many ways of doing this, and
it is unclear which is best.  Should the generated text simply be dumped into
a conventional word-processor, as done in ICG \cite{springer:iaai}?
Or should users make changes with a structured editor, perhaps using the
`linguistic spreadsheet' idea proposed by \cite{kempen:esprit86}?
Or is it best to ask
the user to edit a conceptual representation, as done in
FoG \cite{goldberg:ieee94}?
\item[Examples:] In many cases, adding examples to a text can greatly increase
its usefulness.
But this is another research topic that has been barely
scratched; Mittal's thesis \cite{mittal:thesis} and subsequent work is the
only research in this area that I am aware of.
What principles govern the creation of good examples, and how can
a system generate examples that not only communicate the target information,
but also are consistent with the user's general world knowledge?
\item[Knowledge acquisition:] This is not really an `NLG' topic, but it is
very important to the success of NLG systems, which usually are
knowledge intensive.  Anything that makes it easier to build knowledge bases
will probably make it easier to build NLG systems.
\end{description}

\newpage
\section{Conclusion}

As NLG technology begins to move out of the lab and into real applications,
the NLG community needs to begin thinking not just about how to generally
improve our understanding of this research area, but also about questions
such as (a) what advantages NLG
offers over simpler approaches; (b) under what circumstances
using NLG `adds value' to real-world systems; and (c) where further advances
in NLG could really increase the usefulness of applied NLG systems.
It will probably be many years before we can confidently provide answers
to these questions, but an important step on this path would be to start
more explicitly discussing and exploring these issues within
the community; I can only hope that the presentation in this paper will at
least in a small way encourage people to start thinking more
about these issues.


\end{document}